\documentclass[final,12pt]{colt2024} 

\title[Bridging the Gap: Rademacher Complexity in Robust and Standard Generalization]{Bridging the Gap: Rademacher Complexity in Robust and Standard Generalization}
\usepackage{times}
\usepackage{bbding}
\usepackage{booktabs}
\usepackage{dsfont}
\usepackage{multirow}

\coltauthor{%
 \Name{Jiancong Xiao} \Email{jcxiao@upenn.edu}\\
 \addr University of Pennsylvania
 \AND
 \Name{Ruoyu Sun} \Email{sunruoyu@cuhk.edu.cn}\\
 \addr The Chinese University of Hong Kong, Shenzhen
 \AND
  \Name{Qi Long} \Email{qlong@upenn.edu}\\
 \addr University of Pennsylvania
 \AND
  \Name{Weijie J. Su} \Email{suw@wharton.upenn.edu}\\
 \addr University of Pennsylvania
}

\begin{document}
\maketitle
\begin{abstract}
Training Deep Neural Networks (DNNs) with adversarial examples often results in poor generalization to test-time adversarial data. This paper investigates this issue, known as adversarially robust generalization, through the lens of Rademacher complexity. Building upon the studies by \citet{khim2018adversarial,yin2019rademacher}, numerous works have been dedicated to this problem, yet achieving a satisfactory bound remains an elusive goal. Existing works on DNNs either apply to a surrogate loss instead of the robust loss or yield bounds that are notably looser compared to their standard counterparts. In the latter case, the bounds have a higher dependency on the width $m$ of the DNNs or the dimension $d$ of the data, with an extra factor of at least $\mathcal{O}(\sqrt{m})$ or $\mathcal{O}(\sqrt{d})$. 

This paper presents upper bounds for adversarial Rademacher complexity of DNNs that match the best-known upper bounds in standard settings, as established in the work of \cite{bartlett2017spectrally}, with the dependency on width and dimension being $\mathcal{O}(\ln(dm))$. The central challenge addressed is calculating the covering number of adversarial function classes. We aim to construct a new cover that possesses two properties: 1) compatibility with adversarial examples, and 2) precision comparable to covers used in standard settings. To this end, we introduce a new variant of covering number called the \emph{uniform covering number}, specifically designed and proven to reconcile these two properties. Consequently, our method effectively bridges the gap between Rademacher complexity in robust and standard generalization.\footnote{Accepted for presentation at the Conference on Learning Theory (COLT) 2024.}
\end{abstract}


\section{Introduction}
Deep neural networks (DNNs) are often highly susceptible to adversarial perturbations that are imperceptible to the human eye \citep{goodfellow2015explaining,madry2018towards}. This vulnerability has received significant attention in the machine learning literature over recent years, and a large number of defense algorithms have been proposed to improve robustness in practice \citep{gowal2020uncovering,rebuffi2021fixing}. Nonetheless, these methods still fail to deliver satisfactory performance. One major challenge stems from adversarially robust generalization: DNNs trained with adversarial examples often struggle to generalize well to test-time adversarial data.

\begin{table}[htbp]\small
    \centering
    \caption{List of robust generalization analyses via ARC. Type {\uppercase\expandafter{\romannumeral1}} analysis cannot be applied to DNNs. Type {\uppercase\expandafter{\romannumeral2}} analysis is performed for surrogate losses rather than the robust loss. Type {\uppercase\expandafter{\romannumeral3}} analysis yields looser bounds compared to their standard counterparts. Here, LP and CN stand for layer peeling and covering number, respectively.}\begin{tabular}{clcccc}
    \toprule
  \multirow{2}{*}{\textbf{Type}} & &\multirow{2}{*}{\textbf{Methods}}& \multicolumn{3}{c}{\textbf{Impossible Trinity?}}\\
    \cmidrule(rl){4-6}
        &  & & DNNs & Robust Loss  & Matching \\
    \midrule
   \multirow{3}{*}{\uppercase\expandafter{\romannumeral1}} & \cite{khim2018adversarial} (Thm. 1) &Optimal Attack & 1-Layer & \Checkmark & \Checkmark \\
   &  \cite{yin2019rademacher} (Thm. 1) &Optimal Attack   & 1-Layer & 
   \Checkmark &  \Checkmark \\
   &\cite{awasthi2020adversarial} (Thm. 4) & Optimal Attack & 1-Layer & \Checkmark & \Checkmark \\
      \midrule
   \multirow{3}{*}{\uppercase\expandafter{\romannumeral2}} & \cite{khim2018adversarial} (Thm. 2) &LP + Surrogate Loss & \Checkmark & Tree-Loss & \Checkmark \\
   &   \cite{yin2019rademacher} (Thm. 8) & CN + Surrogate Loss  & 2-Layer & SDP-Loss & \Checkmark \\
& \cite{gao2021theoretical} &CN + Surrogate Loss & \Checkmark & FGSM-Loss & - \\
\midrule
   \multirow{3}{*}{\uppercase\expandafter{\romannumeral3}} &\cite{awasthi2020adversarial}& CN + Optimal Attack & 2-Layer & \Checkmark & \XSolidBrush \\
&\cite{xiao2022adversarial} & CN on weight space & \Checkmark & \Checkmark & \XSolidBrush \\
&\cite{mustafa2022generalization} & CN on perturbation set & \Checkmark & \Checkmark & \XSolidBrush \\
\midrule
&\textbf{Ours} & \textbf{Uniform CN} & \Checkmark & \Checkmark & \Checkmark\\
\bottomrule
    \end{tabular}
    \label{tab:my_label}
\end{table}

In classical learning theory, it is well-known that the generalization gap can be bounded by the Rademacher complexity \citep{bartlett1998sample}. A useful starting point is to consider linear predictors $f:x\rightarrow w^\top x$, which map the input $x$ to the label $y$. For this class, the generalization gap (with respect to Lipschitz losses $\ell(\cdot,\cdot)$), given $n$ training examples with norms bounded by $B$, scales as $\mathcal{O}(B|w|/\sqrt{n})$. To further explore the generalization of deep learning, a series of works aimed at providing better Rademacher complexity bounds for DNNs \citep{bartlett2002rademacher,neyshabur2015norm,golowich2018size}, mainly using tools of layer peeling and covering number. Covering numbers and Rademacher complexities are, in some usual settings, nearly tight with each other \citep{telgarsky2021deep}; however, in this paper, we will only focus on upper bounding Rademacher complexity with covering numbers. We refer to the approaches for bounding standard Rademacher complexity as \emph{standard approaches}. The tightest bound is given by \citet{bartlett2017spectrally}. Since then, progress in the field of norm-based bounds for standard training has experienced a temporary stall, with no tighter bounds being proposed in recent years.

To study the issue of adversarially robust generalization, \citet{khim2018adversarial} and \citet{yin2019rademacher} concurrently extended Rademacher complexity to adversarial settings. They showed that the robust generalization gap can be bounded by \emph{adversarial Rademacher complexity (ARC)}, which is defined by replacing the standard loss function $\ell(f(x),y)$ in Rademacher complexity with the adversarially robust loss $\max_{\|x-x'\|\leq\varepsilon}\ell(f(x'),y)$, where $\varepsilon$ is the attack intensity. However, providing a satisfactory bound for ARC remains an unresolved challenge in the field. Existing research has shown that harmonizing DNNs, robust loss, and a bound that matches its corresponding standard bound appears to represent an impossible trinity, as listed in Table \ref{tab:my_label}. Below, we provide the details.
\paragraph{Type (\uppercase\expandafter{\romannumeral1}): Robust Loss and Matching Bounds.} To avoid the conflict between standard approaches and the $\max$ operation, the ideal way is to find closed-form solutions for optimal attacks $x^*=\arg\max_{\|x-x'\|\leq\varepsilon}\ell(f(x'),y)$. Then, it is able to apply standard approaches to $\ell(f(x^*),y)$. Using this method, \cite{khim2018adversarial} and \citet{yin2019rademacher} provided bounds for ARC in linear functions, and \citet{awasthi2020adversarial} further improved the linear bounds. Nonetheless, finding closed-form solutions for optimal attacks in the context of DNNs is exceedingly complex. Consequently, generalizing this approach to DNNs remains a significant challenge.
\paragraph{Type (\uppercase\expandafter{\romannumeral2}): DNNs and Matching Bounds.} To apply standard approaches to DNNs, several surrogate losses $\hat{\ell}(f(x),y)\approx\max_{\|x-x'\|\leq\varepsilon}\ell(f(x'),y)$ have been designed, where the surrogate losses do not contain a $\max$ operation. These include tree-transformation loss \citep{khim2018adversarial}, SDP relaxation loss \citep{yin2019rademacher}, and FGSM loss \citep{gao2021theoretical}. However, this approach provides upper bounds for Rademacher complexity on surrogate losses rather than the actual adversarially robust loss. Thus, it cannot provide a bound for ARC or the robust generalization gap.
\paragraph{Type (\uppercase\expandafter{\romannumeral3}): DNNs and Robust Loss.} \cite{awasthi2020adversarial} explored solutions for optimal attacks in two-layer neural networks and provided two bounds: one (cf. Thm 7) for a general assumption, but with an extra factor of $\mathcal{O}(\sqrt{m})$, and the other one (cf. Thm 10) is width-independent but requires additional assumptions. In our earlier work \citep{xiao2022adversarial}, we provide the first bound for ARC of DNNs. After that, \cite{mustafa2022generalization} introduced a different bound. These two bounds are obtained by calculating the covering number of adversarial function classes: one based on the weight space and the other on the perturbation set. However, they exhibit a higher dependency on the width of the DNNs and the dimension of the data, respectively. Further discussion will be provided later. We refer to the approaches for bounding ARC as \emph{adversarial approaches}.

Due to the suboptimal nature of existing adversarial bounds, they are inadequate for comprehending robust generalization. Bridging the gap between Rademacher complexity in robust and standard generalization is crucial for gaining a deeper understanding of robust generalization. In this paper, we presents upper bounds for adversarial Rademacher complexity of DNNs that match
the best-known upper bounds in standard settings, as established in the work of \cite{bartlett2017spectrally}. This provides a new insight on understanding robust generalization: the complexity of standard and robust generalization is nearly identical.

\subsection{Main Result}
To state the bound, some notation is necessary. The notation mainly follows the work of \citet{bartlett2017spectrally}. The networks will use $L$ fixed activation functions $(\sigma_1,\cdots,\sigma_{L})$, where $\sigma_i$ is $\rho_i$-Lipschitz and $\sigma_i(0) = 0$. Let $\ell(\cdot,y)$ be a $\rho$-Lipshitz function with respect to the first argument and  takes values in $[0, 1]$. Given $L$ weight matrices $W = (W_1,\cdots, W_L)$ with $W_l\in\mathbb{R}^{m_l\times m_{l-1}}$, let the deep neural networks be $f(x)=\sigma_L W_L\sigma_{L-1}(W_{L-1}\cdots \sigma_1(W_1x)\cdots)$. The network output $f(x)\in \mathbb{R}^{m_L}$ (with $m_0 = d$ and $m_L = k$) is converted to a class label in $\{1,\cdots,k\}$ by taking the $\arg\max$ over components, with an arbitrary rule for breaking ties. Whenever input data $x_1,\cdots, x_n \in \mathbb{R}^d$ are given with $\|x_i\|_2\leq B$, collect them as columns of a matrix $X \in \mathbb{R}^{d\times n}$. Let $\mathcal{B}(x)$ be arbitrary perturbation set around $x$. For example, for $\ell_p$ attack, we denote $\mathcal{B}_{\varepsilon}^p(x)=\{x'\mid\|x-x'\|_p\leq\varepsilon\}$. Let $\gamma$ be the margin. The $\ell_p$ norm $\|\cdot\|_p$ is always computed entry-wise. Thus, for a matrix, $\|\cdot\|_2$ corresponds to the Frobenius norm. Finally, let $\|\cdot\|_\sigma$ denote the spectral norm.
\begin{theorem}
\label{thm:main}
    Let nonlinearities $(\sigma_1,\cdots,\sigma_{L})$ be given as above. Let the network $f: \mathbb{R}^d \rightarrow \mathbb{R}^k$ with weight matrices $W = (W_1,\cdots, W_L)$ have spectral norm bounds $(s_1,\cdots, s_L)$ and $\ell_1$-norm bounds $(a_1, \cdots , a_L)$. Then for $\mathcal{S}=\{(x_i, y_i)\}_{i=1}^n$ drawn i.i.d. from any probability distribution $\mathcal{D}$ over $\mathbb{R}^d \times \{1, \cdots, k\}$, with probability at least $1- \delta$ over $\mathcal{S}$, the adversarially robust generalization gap satisfies
{\footnotesize\begin{equation*}\label{stdbound}\mathbb{E}_\mathcal{D}\max_{x'\in \mathcal{B}(x)}\ell(f(x'),y)-\mathbb{E}_\mathcal{S}\max_{x'\in \mathcal{B}(x)}\ell(f(x'),y)\leq
\tilde{\mathcal{O}}\bigg(\frac{\tilde{B}\rho\prod_{i=1}^L\rho_i s_i}{\sqrt{n}}\bigg(\sum_{i=1}^L\frac{a_i^{2/3}}{s_i^{2/3}}\bigg)^{3/2}+\sqrt{\frac{\ln(1/\delta)}{n}}\bigg),
\end{equation*}}where $\tilde{B}$ is the magnitude of adversarial examples, i.e., $\|x'\|_2\leq \tilde{B}$, $\forall x'\in\mathcal{B}(x)$ and $x\in\{x_i\}_{i=1}^n$.
\end{theorem}
Theorem \ref{thm:main} not only improves upon the work referenced in Table \ref{tab:my_label} but also matches the standard bounds: By replacing $\tilde{B}$ with $B$, the upper bound in Theorem \ref{thm:main} becomes the standard bound established in \cite{bartlett2017spectrally}, v1. The presence of $\tilde B$ is necessary in adversarial settings, as discussed in \cite{yin2019rademacher}.

\paragraph{Margin Bounds.} By replacing $\rho$ by $1/\gamma$ in the right-hand side of the upper bound in Theorem \ref{thm:main}, we obtain a margin bounds for robust generalization gap of the perdition error defined as
{\footnotesize\begin{equation*}
    \begin{aligned}
& \mathbb{P}_{(x, y)\sim\mathcal{D}} \left\{ \exists~x' \in \mathcal{B}(x)~\text{s.t.}~ y \neq \arg\max_{y' \in [k]} f(x')_{y'} \right\} 
-  \frac{1}{n}\sum_{i=1}^n \mathds{1}\bigg( \exists~x_i' \in \mathcal{B}(x_i)~\text{s.t.}~ f( x_i' )_{y_i} \le \gamma + \max_{y' \neq y_i}f( x_i' )_{y'}\bigg).
\end{aligned}
\end{equation*}}
\paragraph{Magnitude of Adversarial Examples.} The form of $\tilde{B}$ depends on $\mathcal{B}(x)$. For $\ell_p$ attacks, i.e., $\|x-x'\|_p\leq \varepsilon$, we have $\tilde B\leq B +\max\{1,d^{\frac{1}{2}-\frac{1}{p}}\}\varepsilon$. There exists an additional dependency on the dimension-$d$ within the magnitude $\tilde B$. It arises from the discrepancy between $\ell_p$ attacks and the $\ell_2$-norm of training samples. We defer the detailed discussion to Section \ref{Sec:gen}.


\subsection{Technical Overview} 
As previously stated, the generalization gap is upper bounded by the covering number of the function class. For simplicity, we refer to the cover and covering number of the adversarial function class as the adversarial cover and adversarial covering number, respectively. This paper primarily addresses the question: How can we estimate the adversarial covering number? We briefly discuss why it is challenging to extend two related lines of work to provide a strong estimate of the adversarial covering number.

1) A key component in standard approaches is the Maurey sparsification lemma, which provides a strong estimate of the covering number of the matrix product $Wx$, where $W$ is the weight of a layer and $x$ is the input of the corresponding layer. In the adversarial setting, $x$ is dependent on $W$ and the weights of deeper layers. It is unclear how the Maurey sparsification lemma can be applied to this setting.

2) To ensure compatibility with adversarial examples, alternative approaches have been devised in adversarial settings that do not rely on the matrix product $Wx$ and thus do not use the Maurey sparsification lemma. These approaches lead to bounds on the adversarial covering number with a higher dependency on the width $m$ or the data dimension $d$.

To remove the extra factors in the existing bounds of the adversarial covering number, we suspect that the Maurey sparsification lemma is still needed. Thus, we set up the following goal: Can we devise an approach that is amenable to the Maurey sparsification lemma and is compatible with adversarial examples? We describe our approach to achieve this goal.

Firstly, we propose a new variant of the covering number called the \emph{uniform covering number}. Let $\mathcal{W}$ be a weight matrix space. Consider the matrix product $Wx'$, where $x'\in\mathcal{B}(x)$ and $W\in\mathcal{W}$. Informally, we say a subset $\mathcal{C}$ is a uniform cover of $\mathcal{W}$ with respect to $\mathcal{B}(x)$, if for all $x'\in\mathcal{B}(x)$, ${W' x':W'\in\mathcal{C}}$ is always a cover of ${W x':W\in\mathcal{W}}$.

This definition retains the matrix product form $Wx'$, preserving the potential for utilizing the Maurey sparsification lemma. Meanwhile, this variant of cover is designed to be uniform across the perturbation set $\mathcal{B}(x)$ of the corresponding layer, regardless of $x$'s reliance on weights from deeper layers. This design ensures its potential compatibility with adversarial examples.
The following task is to prove that a precise adversarial cover can indeed be established using the uniform cover for the weight matrix space.

Next, we inductively determine the uniform cover $\mathcal{C}_i$ of each layer, $i=1,\cdots,L$.
Then, we consider the function class parameterized by weight matrices in $\mathcal{C}_L\times\cdots\times\mathcal{C}_1$, which constitutes a subset of the adversarial function class.
To prove that this is an adversarial cover, the following lemma is required: Informally, the distance between any two adversarial functions can be bounded by the distance between two standard functions evaluated at an intermediate adversarial example, which is proposed in our earlier work \citep{xiao2022adversarial} and will be introduced later.

By the definition of uniform cover, for all $x'$, ${W'x'}$ is always a cover of ${Wx'}$. This remains valid for any determined intermediate adversarial examples, regardless of their dependency on different pairs of adversarial functions. Consequently, this allows us to inductively prove that the uniform cover of each layer constitutes an adversarial cover.

Finally, we bound the uniform covering number using the Maurey sparsification lemma, ensuring that such an adversarial cover is as precise as a standard cover. This approach results in a bound that is not only tighter than existing adversarial bounds but also matches the bounds in standard settings.

\section{Covering in Standard Settings and Main Challenge in Adversarial Settings}
\label{Sec:challenge}
\subsection{Preliminaries}
\paragraph{Function Class.} We consider the function class (or hypothesis class) of neural networks as follow:
\begin{equation*}
\label{hypo}
    \mathcal{H}=\{h:(x,y)\rightarrow \ell(\sigma_L W_L\sigma_{L-1}(W_{L-1}\cdots \sigma_1(W_1x)\cdots),y)\mid W_i\in \mathcal{W}_i, i=1,\cdots,L\}.
\end{equation*}



\paragraph{Adversarial Function Class.} The adversarial function class of neural networks is defined as follow:
\begin{equation}
\label{advhypo}
\begin{aligned}
    \tilde{\mathcal{H}}
    =&\{\tilde{h}:(x,y)\rightarrow \max_{x'\in\mathcal{B}(x)}h(x,y)\mid h\in \mathcal{H}\},
\end{aligned}
\end{equation}
Given a dataset $\mathcal{S}=\{(x_i,y_i)\}_{i=1}^n$, the function classes $\mathcal{H}$ and $\tilde{\mathcal{H}}$ on $\mathcal{S}$ are defined as $\mathcal{H}_{|\mathcal{S}}=\{(h(x_1,y_1),\cdots,h(x_n,y_n))\mid h\in\mathcal{H}\}$ and $\tilde{\mathcal{H}}_{|\mathcal{S}}=\{(\tilde{h}(x_1,y_1),\cdots,\tilde{h}(x_n,y_n))\mid\tilde{h}\in\tilde{\mathcal{H}}\}$, respectively.
\begin{definition}[Covering Number]
	\label{def:cov}
	Let $\epsilon>0$ and $(\mathcal{W}$,$\|\cdot\|)$ be a normed space. We say $\mathcal{C}\subset \mathcal{W}$ is an $\epsilon$-cover\footnote{We use two different Greek alphabet: $\varepsilon$ for adversarial attacks and $\epsilon$ for covering number.} of $\mathcal{W}$, if for any $W\in \mathcal{W}$, there exists $W'\in\mathcal{C}$ s.t. $\|W-W'\|\leq\epsilon$. The
least cardinality of such subset $\mathcal{C}$ is called the $\epsilon$-covering number, denoted as $\mathcal{N}(\mathcal{W},\epsilon, \|\cdot\|)$.
\end{definition}
We use $\mathcal{N}(\mathcal{W})$ as an abbreviation to denote the covering number of $\mathcal{W}$ when it does not cause ambiguity. Since the main problem of this paper is how to compute the adversarial covering number $\mathcal{N}(\tilde{\mathcal{H}})$, we will first focus on this problem in Section \ref{Sec:challenge} and \ref{Sec:cover}. As for the other preliminaries such as the definition of robust generalization and how to covert adversarial covering number to robust generalization, they follow existing work. We leave them to Section \ref{Sec:gen}, where we complete the proof of Theorem \ref{thm:main}.

\subsection{Covering Number Bounds in Standard Settings}
A well-known upper bound in standard settings is provided by \cite{bartlett2017spectrally}. We first give a brief review of \cite{bartlett2017spectrally}'s approach.
\paragraph{Step (\uppercase\expandafter{\romannumeral1}): Matrix Covering.} The most important building block is the matrix covering of the affine transformation $WX$, where $W$ is the weight matrix, and $X$ is the data passed through the network. Denote the ($q,s$)-group norm $\|W\|_{q,s}$ as the $q$-norm of the $s$-norm of the rows of $W$.

\begin{lemma}[\cite{bartlett2017spectrally}, Lemma 3.2; \cite{zhang2002covering}, Theorem 3]
\label{lemma:zhang}
Let conjugate expone-nts $(p, q)$ and $(r, s)$ be given with $p \leq 2$, as well as positive reals $(a, b, \epsilon)$ and positive integer $m$. Let matrix $X \in \mathbb{R}^{d\times n}$ be given with $\|X\|_p \leq b$. Then
\begin{equation*}
    \ln \mathcal{N}\bigg(\bigg\{WX:W\in\mathbb{R}^{m\times d},\|W^\top\|_{q,s} \leq a\bigg\},\epsilon,\|\cdot\|_2\bigg)\leq\bigg \lceil\frac{a^2b^2m^{\frac{2}{r}}}{\epsilon^2}\bigg\rceil \ln(2dm).
\end{equation*}
\end{lemma}
The proof utilizes Maurey sparsification lemma \citep{pisier1981remarques}.

\paragraph{Step (\uppercase\expandafter{\romannumeral2}): Induction on Layers.} 
Denote $X_{i}$ as the (fixed) output of the $i^{th}$ layer. It is proven by induction that the covering number of the whole neural network function class $\mathcal{H}$ is bounded by the sum of the matrix covering number of the output spaces of each of the $i^{th}$ layers. 
\begin{equation}
\label{eq:stddecomp}
    \ln \mathcal{N}(\mathcal{H}_{|\mathcal{S}},\epsilon,\|\cdot\|_2)\leq \sum_{i=1}^L\sup_{(W_1,\cdots,W_{i-1})}\ln \mathcal{N}(\{W_iX_{i-1}:\|W_i^\top\|_{q,s} \leq a_i\},\epsilon_i,\|\cdot\|_2).
\end{equation}
\paragraph{Step (\uppercase\expandafter{\romannumeral3}): Dudley's Integral.} Using the standard Dudley entropy integral, Rademacher complexity is upper bounded by the covering number of the function classes (see e.g. \citet{mohri2018foundations}).
\subsection{Main Challenge in Adversarial Settings}
As discussed in the Introduction, standard approaches, including the approach of \cite{bartlett2017spectrally}, cannot directly utilize the $\max$ operation in robust loss. Specifically, within this covering number approach, only Step (\uppercase\expandafter{\romannumeral3}) can be directly applied to adversarial settings, namely ARC can be bounded by the adversarial covering number via Dudley's integral. The interaction between $X$ and $W$ affects the application of the first two steps to adversarial settings. We refer to the challenge of applying Step (\uppercase\expandafter{\romannumeral1}) and Step (\uppercase\expandafter{\romannumeral2}) as Challenge (\uppercase\expandafter{\romannumeral1}) and Challenge (\uppercase\expandafter{\romannumeral2}), respectively.
\paragraph{Challenge (\uppercase\expandafter{\romannumeral1}).} Denote $\mathcal{B}(X)={[x_1',\cdots,x_n']: x_i'\in \mathcal{B}(x_i), i=1,\cdots,n}$. The function $W\rightarrow X^{adv}(W)$ can be written as $X^{adv}(W)=\arg\max_{X'\in \mathcal{B}(X)} \ell(WX',Y)$. Then, it is unclear how to calculate the covering number of the matrix ${WX^{adv}(W)}$ through Lemma \ref{lemma:zhang} or other related approaches.
\paragraph{Challenge (\uppercase\expandafter{\romannumeral2}).} Given that adversarial examples are not static, constructing a cover for adversarial function classes inductively, as outlined in Step (\uppercase\expandafter{\romannumeral2}), is infeasible. Consequently, the inequality and analogous formulations presented in Step (\uppercase\expandafter{\romannumeral2}) are not applicable in adversarial settings.

\subsection{Existing Adversarial Approaches for the Challenges}
\label{sec:exist}
The challenges mentioned were initially highlighted by \citet*{yin2019rademacher} in their attempt to extend the standard bound \citep{bartlett2017spectrally} to adversarial settings. As a compromise, they employed an SDP-relaxation loss as a surrogate for the adversarially robust loss and applied \citet{bartlett2017spectrally}'s bound to this SDP-relaxation loss. That work did not provide a bound for the original robust loss on DNNs.

The first bound that applies to the robust loss of DNNs was given by our earlier work \citep{xiao2022adversarial}. We used a robustified weight perturbation bound to prove the following decomposition:
\begin{equation*}
    \ln\mathcal{N}(\tilde{\mathcal{H}}_{|\mathcal{S}},\epsilon,\|\cdot\|_2)\leq\sum_{i=1}^L\ln \mathcal{N}(\mathcal{W}_i,\epsilon_i,\|\cdot\|_{\text{op}}).
\end{equation*}
Here, the right-hand side is the covering number of the weight spaces $\mathcal{W}_i$ of each layer, rather than the output space of $\{\mathcal{W}_i\times X_{i-1}\}$. $\|\cdot\|_{\text{op}}$ represents the operator norm.
This bound effectively eliminates the effect of the interaction between $W$ and $X$. However, the weight covering $\mathcal{N}(\mathcal{W})$ cannot be bounded by the Maurey sparsification lemma.

\cite{mustafa2022generalization}'s idea is to take the covering number of the perturbation set $\mathcal{B}(x)$ into account. Firstly, they considered a cover $\mathcal{C}$ for $\mathcal{B}(0)$ and defined an extended dataset $\hat{\mathcal{S}}=\{(x_i+\delta,y_i),i\in [n], \delta\in\mathcal{C}\}$. Secondly, they considered the extended function class $\hat{\mathcal{H}}=\{(x,y,\delta)\rightarrow h(x+\delta,y),h\in\mathcal{H}\}$. Finally, they showed that the covering number of the adversarial function class $\tilde{\mathcal{H}}$ can be bounded by the covering number of $\hat{\mathcal{H}}$, \emph{i.e.,}
\begin{equation*}
    \ln\mathcal{N}(\tilde{\mathcal{H}}_{|\mathcal{S}},\epsilon,\|\cdot\|_2)\leq\ln \mathcal{N}(\hat{\mathcal{H}}_{|\hat{\mathcal{S}}},\frac{\epsilon}{2},\|\cdot\|_2).
\end{equation*}
Since $\hat{{\mathcal{H}}}$ does not contain the $\max$ operation, the covering number on the right-hand side can be bounded through the use of existing standard approaches. However, this approach necessitates the determination of the Lipschitz constant for the function ($\delta\rightarrow h(x+\delta,y)$).

As a result, these two methods result in suboptimal bounds.
\section{Covering Number of Adversarial Function Classes}
\label{Sec:cover}
\citet{bartlett2017spectrally}'s approach showed that the application of Maurey sparsification Lemma is a key component to obtain a tighter bound, yet this approach seems incompatible with adversarial examples. The approaches introduced in \citet{xiao2022adversarial} and \cite{mustafa2022generalization} are compatible with adversarial examples, yet the constructed covers are not as precise as \citet{bartlett2017spectrally}'s cover. These observations suggest that a matching adversarial bound could potentially be derived by fulfilling both requirements: 1) compatibility with adversarial examples, and 2) the application of Maurey sparsification Lemma. To this end, we introduce the concept of the uniform covering number, designed to harmonize these two objectives.

\subsection{Uniform Covering Number}
\begin{definition}[Uniform Covering Number]
\label{def:UCN}
    Let $\mathcal{C}$ be a subset of $\mathcal{W}$. We say $\mathcal{C}$ is an $\epsilon$-uniform cover of $\mathcal{W}$ with respect to $\mathcal{X}$, if $\forall X\in\mathcal{X}$, $\{\tilde{W}X:\tilde{W}\in\mathcal{C}\}$ is always an $\epsilon$-cover of $\{WX:W\in\mathcal{W}\}$ with norm $\|\cdot\|$. The least cardinality of such subset $\mathcal{C}\subset\mathcal{W}$ is called the uniform covering number, denoted as
    \begin{equation*}
        \mathcal{UN}_{\mathcal{X}}(\mathcal{W},\epsilon,\|\cdot\|).
    \end{equation*}
\end{definition}

Our first observation is that the dependency of adversarial examples on deeper layers conflicts with the definition of cover in Lemma \ref{lemma:zhang}. This discrepancy makes it challenging to determine how to effectively bound the covering number of adversarial function classes in relation to the covering number of individual layers. This dilemma has led us to propose a new kind of covering number that is uniformly applicable across perturbation sets. Then, the `new cover' of individual layers is independent to the adversarial examples and to the weights from deeper layers. Uniform covering number is such a new concept by choosing $\mathcal{X}_i$ to be the pertubation set of each layer.  We will show how to build a cover of adversarial function classes based on the uniform covers of each layers.

Secondly, based on the matrix product form in the definition of uniform covering number, we bound uniform covering number using Maurey sparsification Lemma. Thus, uniform covering number serves as a bridge between the covering number of adversarial function classes and the upper bound by Maurey sparsification Lemma. Consequently, it becomes possible to derive a matching upper bound for the covering number of adversarial function classes.

By the definition of uniform covering number. We directly have $\mathcal{N}(\{WX:W\in\mathcal{W}\},\epsilon,\|\cdot\|)\leq\mathcal{UN}_{\mathcal{X}}(\mathcal{W},\epsilon,\|\cdot\|)$, for all $X\in\mathcal{X}$. Therefore, we have 
$$\sup_{X\in\mathcal{X}}\mathcal{N}(\{WX:W\in\mathcal{W}\},\epsilon,\|\cdot\|)\leq\mathcal{UN}_{\mathcal{X}}(\mathcal{W},\epsilon,\|\cdot\|).$$
In general, the above equality does not hold. The uniform covering number cannot be readily simplified or directly expressed in terms of the covering number. Furthermore, the left-hand side is used to bound the covering number of standard function classes, as presented in Eq. (\ref{eq:stddecomp}), yet its application for bounding the covering number of adversarial function classes remains ambiguous. This distinction necessitates the introduction of a new definition.

Finally, we present the subsequent Lemma, crucial for demonstrating that a uniform cover is capable of constituting a cover for adversarial function classes.
\begin{lemma}[Intermediate Adversarial Example \citep{xiao2022adversarial}]
\label{lemma:iae}Given $(x,y)$ and perturbation set $\mathcal{B}(x)$. For all $\tilde{h}_1,\tilde{h}_2\in\tilde{\mathcal{H}}$ with their standard counterparts $h_1,h_2\in\mathcal{H}$, there exists an adversarial example $x'(\tilde{h}_1,\tilde{h}_2)\in \mathcal{B}(x)$, s.t.
\begin{equation*}
    |\tilde{h}_1(x,y)-\tilde{h}_2(x,y)|\leq|h_1(x'(\tilde{h}_1,\tilde{h}_2),y)-h_2(x'(\tilde{h}_1,\tilde{h}_2),y)|.
\end{equation*}
We refer to this adversarial example $x'(\tilde{h}_1,\tilde{h}_2)\in \mathcal{B}(x)$ as intermediate adversarial example.
\end{lemma}
Lemma \ref{lemma:iae} plays a crucial role in bounding adversarial functions with standard functions, eliminating the $\max$ operation to enable mathematical induction across layers. Notably, $x'(\tilde{h}_1,\tilde{h}_2)$ varies based on $\tilde{h}_1, \tilde{h}_2$, and their weights across all layers. Our concept of the uniform covering number is designed to accommodate this dependency effectively.

\subsection{Proof Sketch of Covering Number Bounds of Adversarial Function Class}
\paragraph{Step (\uppercase\expandafter{\romannumeral1}): Uniform Covering.} Our first step is to bound the uniform covering number via Maurey sparsification Lemma.
\begin{lemma}[Upper Bounds of Uniform Covering Number]
\label{lemma:unupperbound}
Given positive reals $(a, b, \epsilon)$ and positive integer $(d,m)$. Let $\|X\|_2 \leq b$, for all $X\in\mathcal{X}$. Let $\|W\|_1\leq a$ for all $W\in\mathcal{W}$. Then
\begin{equation*}
\ln \mathcal{UN}_{\mathcal{X}}(\mathcal{W},\epsilon,\|\cdot\|) \leq\bigg\lceil\frac{a^2b^2}{\epsilon^2}\bigg\rceil \ln(2dm).
\end{equation*}
\end{lemma}
The application of Maurey sparsification Lemma leads to a $\mathcal{O}(\ln(dm))$ dependency on width and dimension. Absorbing the logarithmic factors, the dependency is $\tilde{\mathcal{O}}(1)$, as presented in Theorem \ref{thm:main}. In the first and second versions of \citet{bartlett2017spectrally}, two bounds are discussed: the first one focuses on the $1$-norm of the weight matrix $W$, while the second one is based on the $2,1$-norm of $W$, with other factors remaining consistent between the two. The proof for the $2,1$-norm bound constructs a cover that depends on the data by normalizing each row of $X$ by the norms of its respective rows. Consequently, the $2,1$-norm bound is data-dependent, and its applicability is limited in adversarial contexts due to its reliance on the data $X$. However, this distinction is relatively minor, as the two norms are close. Crucially, the key insight of reducing dependency on $m$ and $d$ to $\ln(dm)$ remains applicable in adversarial settings.
\paragraph{Step (\uppercase\expandafter{\romannumeral2}): Induction on Layers.} The next lemma shows that the covering number of adversarial classes can be bounded in terms of the uniform covering number.
\begin{lemma}[Covering of Adversarial Function Classes]
\label{lemma:coverAFC}
    Let $(\epsilon_1,\cdots,\epsilon_L)$ be given, along with Lipschitz activation function $\sigma_i$ (where $\sigma_i(\cdot)$ is $\rho_i$-Lipschitz, $i=1,\cdots,L$), fixed Lipschitz loss function $\ell$ ($\ell$ is $\rho$-Lipschitz) and operator norm bounds $(c_1,\cdots, c_L)$. Suppose the matrices $W =(W_1,\cdots, W_L)$ lie within $\mathcal{W}_1 \times \cdots\times \mathcal{W}_L$ where $\mathcal{W}_i$ are arbitrary classes with the property that each $W_i \in \mathcal{W}_i$ has $\|W_i\|_{op}\leq c_i$. Starting from the pertubation set, let $\mathcal{X}_0=\mathcal{B}(X)$). For $i=1,\cdots,L-1$, let $\mathcal{X}_i=\{\sigma_i(W_{i}X_{i-1}):W_i\in\mathcal{W}_i, X_{i-1}\in\mathcal{X}_{i-1}\}$. Then, letting $\epsilon=\rho \sum_{j\leq L} \epsilon_j\rho_j \prod_{l=j+1}^L \rho_l c_l$, the adversarial function class $\tilde{\mathcal{H}}$ have covering number bound
    \begin{equation*}
    \ln \mathcal{N}(\tilde{\mathcal{H}}_{|\mathcal{S}},\epsilon,\|\cdot\|_2)\leq\sum_{i=1}^L \ln\mathcal{UN}_{\mathcal{X}_{i-1}}(\mathcal{W}_i,\epsilon_i,\|\cdot\|_2).
    \end{equation*}
\end{lemma}
The foundation of the proof combines the concept of uniform covering number, as outlined in Definition \ref{def:UCN}, with the principle of intermediate adversarial examples introduced in Lemma \ref{lemma:iae}. The proof is provided in Section \ref{sec:proof}. By combining Lemma \ref{lemma:unupperbound} and Lemma \ref{lemma:coverAFC}, we obtain the upper bound for the covering number of adversarial function classes. In this context, $\bar m=\max\{m_0,\cdots,m_L\}$.

\begin{theorem}
\label{thm:main2}
    Let nonlinearities $(\sigma_1,\cdots,\sigma_{L})$ be given, where $\sigma_i$ is $\rho_i$-Lipschitz and $\sigma_i(0) = 0$. Let the loss function $\ell$ be $\rho$-Lipschitz. Let the network $f : \mathbb{R}^d \rightarrow \mathbb{R}^k$ with weight matrices $W = (W_1,\cdots, W_L)$ have spectral norm bounds $(s_1,\cdots, s_L)$, and $\ell_1$-norm bounds $(a_1, \cdots , a_L)$. Then, the adversarial function class $\tilde{\mathcal{H}}_{|\mathcal{S}}$ have covering number bound
\begin{equation*}\label{cnbound}\ln \mathcal{N}(\tilde{\mathcal{H}}_{|\mathcal{S}},\epsilon,\|\cdot\|_2)\leq
\frac{\tilde{B}^2\rho\ln(2\bar m^2)}{\epsilon^2}\bigg(\prod_{i=1}^L\rho_i s_i\bigg)\bigg(\sum_{i=1}^L\frac{a_i^{2/3}}{s_i^{2/3}}\bigg)^{3/2},
\end{equation*}where $\tilde{B}$ is the magnitude of adversarial examples, i.e., $\|x'\|_2\leq \tilde{B}$, $\forall x'\in\mathcal{B}(x)$ and $x\in\{x_i\}_{i=1}^n$.
\end{theorem}
\paragraph{Step (\uppercase\expandafter{\romannumeral3}): Dudley's Integral.} Using the standard Dudley entropy integral, ARC is upper bounded by the covering number of the adversarial function classes.

\subsection{Proof of Lemma \ref{lemma:iae}}
Proof: Let 
$$x(\tilde{h}_1)=\arg\max_{x'\in\mathcal{B}(x)}h_1(x',y),\quad x(\tilde{h}_2)=\arg\max_{x'\in\mathcal{B}(x)}h_2(x',y).$$
Then, 
    \begin{equation*}\small
    \begin{aligned}
    \big|\tilde{h}_1(x,y)-\tilde{h}_2(x,y)\big|\leq \max\big\{|h_1(x(\tilde{h}_1),y)-h_2(x(\tilde{h}_1),y)|,|h_1(x(\tilde{h}_2),y)-h_2(x(\tilde{h}_2),y)|\big\}.
    \end{aligned}     
    \end{equation*}
It is because $$h_1(x(\tilde{h}_1),y)-h_2(x(\tilde{h}_2),y)\leq h_1(x(\tilde{h}_1),y)-h_2(x(\tilde{h}_1),y)$$and
$$h_2(x(\tilde{h}_2),y)-h_1(x(\tilde{h}_1),y)\leq h_2(x(\tilde{h}_2),y)-h_1(x(\tilde{h}_2),y).$$ 
Let
\begin{equation}
\label{eq:iae}
    x'(\tilde{h}_1,\tilde{h}_2)=\begin{cases}
        x(\tilde{h}_1),\ \text{if}\  h_1(x(\tilde{h}_1),y)\geq h_2(x(\tilde{h}_2),y)\\
        x(\tilde{h}_2),\ \text{if}\  h_1(x(\tilde{h}_1),y)<h_2(x(\tilde{h}_2),y).
    \end{cases}
\end{equation}
We have
\begin{equation*}
    |\tilde{h}_1(x,y)-\tilde{h}_2(x,y)|\leq|h_1(x'(\tilde{h}_1,\tilde{h}_2),y)-h_2(x'(\tilde{h}_1,\tilde{h}_2),y)|.
\end{equation*}
The expression in Eq. (\ref{eq:iae}) is the intermediate adversarial examples.

\subsection{Proof of Lemma \ref{lemma:coverAFC}}
\label{sec:proof}
For $i=1,\cdots,L$, let $\mathcal{C}_i$ be an $\epsilon_i$-uniform cover of $\mathcal{W}_i$ with respect to $\mathcal{X}_{i-1}$. This forms a subset of the adversarial function class $\tilde{\mathcal{H}}$, denoted as
\begin{equation*}
\label{coveringset}
    \tilde{\mathcal{C}}=\bigg\{\max_{x'\in\mathcal{B}(x)} \ell(\sigma_L W'_L\sigma_{L-1}(W'_{L-1}\cdots \sigma_1(W'_1x')\cdots),y)\mid W'_i\in \mathcal{C}_i, i=1,\cdots,L\bigg\}.
\end{equation*}
Given $(x,y)$, for all $\tilde{h}_1\in\tilde{\mathcal{H}}$ and $\tilde{h}_2\in\tilde{\mathcal{C}}$, by Lemma \ref{lemma:iae}, there exist an intermediate adversarial examples $x'(\tilde{h}_1,\tilde{h}_2)$, such that
\begin{equation*}
    |\tilde{h}_1(x,y)-\tilde{h}_2(x,y)|\leq|h_1(x'(\tilde{h}_1,\tilde{h}_2),y)-h_2(x'(\tilde{h}_1,\tilde{h}_2),y)|.
\end{equation*}
Given dataset $\mathcal{S}$ with data matrix $X$ and $Y$, denotes $X'(\tilde{h}_1,\tilde{h}_2)\in\mathbb{R}^{d\times n}$ as the collection of intermediate adversarial examples of $X$. Finally, Let $X_{i}(\tilde{h}_1,\tilde{h}_2)$ and $\hat{X}_{i}(\tilde{h}_1,\tilde{h}_2)$ be the output of $X'(\tilde{h}_1,\tilde{h}_2)$ pass through the first to the $(i-1)^{th}$ layer of $\tilde{h}_1$ and $\tilde{h}_2$, i.e.,
$$X_{i}(\tilde{h}_1,\tilde{h}_2)=\sigma_{i-1}(W_{i-1}\cdots \sigma_1(W_1 X'(\tilde{h}_1,\tilde{h}_2))\cdots), i =2,\cdots,L,$$
$$\hat{X}_{i}(\tilde{h}_1,\tilde{h}_2)=\sigma_{i-1}(W'_{i-1}\cdots \sigma_1(W'_1 X'(\tilde{h}_1,\tilde{h}_2))\cdots), i =2,\cdots,L,$$
respectively. Then,
\begin{equation}
\label{eq4}
    \begin{aligned}
        \Delta_{i+1}:&=\|X_{i}(\tilde{h}_1,\tilde{h}_2)-\hat{X}_{i}(\tilde{h}_1,\tilde{h}_2)\|\\&\leq \rho_i\|W_i X_{i-1}(\tilde{h}_1,\tilde{h}_2)-W'_i \hat{X}_{i-1}(\tilde{h}_1,\tilde{h}_2)\|\\
        &\leq\rho_i\big(\|W_i\|_\sigma \Delta_{i}+\|W_i \hat{X}_{i-1}(\tilde{h}_1,\tilde{h}_2)-W'_i \hat{X}_{i-1}(\tilde{h}_1,\tilde{h}_2)\|\big).
    \end{aligned}
\end{equation}
By the definition of $\mathcal{X}_{i-1}$, we have $\hat{X}_{i-1}(\tilde{h}_1,\tilde{h}_2)\in \mathcal{X}_{i-1}$. Since $\mathcal{C}_i$ is a $\epsilon_i$-uniform cover of $\mathcal{W}_i$ with respect to $\mathcal{X}_{i-1}$, it follows that $\{W'_i \hat{X}_{i-1}(\tilde{h}_1,\tilde{h}_2)\mid W'_i\in\mathcal{C}_i\}$ is an $\epsilon_i$-cover for $\{W_i \hat{X}_{i-1}(\tilde{h}_1,\tilde{h}_2)\mid W_i\in\mathcal{W}_i\}$, for all $\tilde{h}_1,\tilde{h}_2$. Then
\begin{equation}
\label{eq5}
    \|W_i \hat{X}_{i-1}(\tilde{h}_1,\tilde{h}_2)-W'_i \hat{X}_{i-1}(\tilde{h}_1,\tilde{h}_2)\|\leq\epsilon_i.
\end{equation}
By combining Eq. (\ref{eq4}) and Eq. (\ref{eq5}), we have
\begin{equation}
\label{eq6}
    \begin{aligned}
    \Delta_{i+1}\leq\rho_i\big(\|W_i\|_\sigma \Delta_{i}+\epsilon_i\big).
    \end{aligned}
\end{equation}
\paragraph{Remark.}
Let $X_i$ denote the output of clean samples $X$ after passing through the layers up to the $(i-1)^{th}$ layer. Substituting $X_i$ for $X_{i}(\tilde{h}_1,\tilde{h}_2)$ in Eq. (\ref{eq6}) reproduces the recursive formulation provided in \citet{bartlett2017spectrally}. The critical distinction between adversarial and standard scenarios lies in the reliance on two functions, $\tilde{h}_1$ and $\tilde{h}_2$. In the standard setting, the training data $X$ remains constant, allowing a standard $\epsilon_i$-cover to facilitate Eq. (\ref{eq6}). This method, however, is not viable when $X$ is dynamic. This dilemma has perplexed the research community for years. Conversely, our concept of a uniform cover efficiently establishes a universally applicable cover for all adversarial examples given $\tilde{h}_1$ and $\tilde{h}_2$. Consequently, our approach addresses the primary issue, demonstrating that the recursive form in Eq. (\ref{eq6}) is applicable in the adversarial setting.

Finally, we complete the proof by mathematical induction. For all $\tilde{h}_1(X,Y)\in\tilde{\mathcal{H}}_{|\mathcal{S}}$ and $\tilde{h}_2(X,Y)\in\tilde{\mathcal{C}}_{|\mathcal{S}}$, using Eq. (\ref{eq6}), we have
\[|\tilde{h}_1(X,Y)-\tilde{h}_2(X,Y)|\leq \rho\Delta_{L+1}\leq\rho \sum_{j\leq L} \epsilon_j\rho_j \prod_{l=j+1}^L \rho_l c_l:=\epsilon.\]
Therefore, $\tilde{\mathcal{C}}_{|\mathcal{S}}$ is an $\epsilon$-cover of the adversarial function class $\tilde{\mathcal{H}}_{|\mathcal{S}}$. We have
    \begin{equation*}
    \ln \mathcal{N}(\tilde{\mathcal{H}}_{|\mathcal{S}},\epsilon,\|\cdot\|_2)\leq \ln|\tilde{\mathcal{C}}_{|\mathcal{S}}|=\sum_{i=1}^L \ln\mathcal{UN}_{\mathcal{X}_{i-1}}(\mathcal{W}_i,\epsilon_i,\|\cdot\|_2).
    \end{equation*}
\section{Adversarial Robust Generalization}
\label{Sec:gen}
In this section, we complete the gap between adversarially robust generalization and covering number, which mainly follows classical learning theory \citep{mohri2018foundations}.
\paragraph{Robust Generalization Gap.} Let the robust population risk and the robust empirical risk be
\[
\Tilde{R}_{\mathcal{D}}(h)=\mathbb{E}_{(x,y)\sim \mathcal{D}}\max_{x'\in\mathcal{B}(x)}h(x',y)\quad and\ \quad
\Tilde{R}_{\mathcal{S}}(h)=\mathbb{E}_{(x,y)\sim \mathcal{S}}\max_{x'\in\mathcal{B}(x)}h(x',y),
\]
respectively. The robust generalization gap is defined as $\Tilde{R}_{\mathcal{D}}(h)-\Tilde{R}_{\mathcal{S}}(h)$.

\begin{definition}[Adversarial Rademacher Complexity]
\label{def:ARC}
Let the Rademacher random variables $\sigma_i$ equals to $1$ and $-1$ with equal probability. ARC is defined by the Rademacher complexity of the adversarial function class $\tilde{\mathcal{H}}$, i.e.  \[\mathcal{R}(\tilde{\mathcal{H}}_{|\mathcal{S}})=\mathbb{E}_\sigma \frac{1}{n}\bigg[\sup_{\tilde{h}\in\tilde{\mathcal{H}}}\sum_{i=1}^{n}\sigma_i \tilde{h}(x,y)\bigg]=\mathbb{E}_\sigma \frac{1}{n}\bigg[\sup_{h\in\mathcal{H}}\sum_{i=1}^{n}\sigma_i \max_{x'\in\mathcal{B}(x)}h(x',y)\bigg].\]
\end{definition}
Then, it is proved that adversarial robust generalization can be bounded by ARC.
\begin{lemma}
[\citet{yin2019rademacher}] 
	\label{prop2} Suppose that the range of the loss function $h(x,y)$ is $[0,1]$. Then, for any $\delta\in(0,1)$, with probability at least $1-\delta$, the following holds for all $h\in\mathcal{H}$,
	\begin{equation*}
	\Tilde{R}_{\mathcal{D}}(h)\leq \Tilde{R}_{\mathcal{S}}(h)+2\mathcal{R}(\mathcal{\Tilde{H}}_{|\mathcal{S}})+3\sqrt{\frac{\log\frac{2}{\delta}}{2n}}.
	\end{equation*}
\end{lemma}
Finally, we introduce the standard Dudley entropy integral bound on the empirical
Rademacher complexity (e.g. \citet{mohri2018foundations}), which is used in the proof of Theorem \ref{thm:main}. 

\begin{lemma}[Dudley's Integral]\label{lemma:dud} Let $\tilde{\mathcal{H}}$ be a real-valued function class taking values in $[0, 1]$, and assume that $0\in\tilde{\mathcal{H}}$. Then
\begin{equation*}
    \mathcal{R}(\mathcal{\Tilde{H}}_{|\mathcal{S}})\leq
    \inf_{\alpha>0}\bigg(\frac{4\alpha}{\sqrt{n}}+\frac{12}{n}\int_\alpha^{\sqrt{n}}\sqrt{\ln \mathcal{N}(\tilde{\mathcal{H}}_{|\mathcal{S}},\epsilon,\|\cdot\|_2)}d\epsilon\bigg).
\end{equation*}
\end{lemma}
Then, by combining Lemma \ref{prop2}, Lemma \ref{lemma:dud} and Lemma \ref{thm:main2}, we obtain our main result of Theorem \ref{thm:main}.

\paragraph{Margin Bounds.} Suppose a neural network computes a function $f : \mathbb{R}^d \rightarrow \mathbb{R}^k$, where $k$ is the number of classes. The most natural way to convert this to a classifier is to select the output coordinate with the largest magnitude, meaning $x \rightarrow \arg\max_j f(x)_j$. The margin, then, measures the gap between the output for the correct label and other labels, defined as $M(f(x),y)=f(x)_y - \max_{j\neq y} f(x)_j$. The function makes a correct prediction if and only if $M(f(x),y)> 0$. $M(f(x),y)$ is $2$-Lipschitz. We consider a particular loss function $\ell (f(x), y) = \phi_\gamma (M(f(x), y))$, where $\gamma > 0$ and $\phi_\gamma:\mathbb{R} \rightarrow [0,1]$ is the ramp loss:
\begin{equation*}\label{eq:ramp}
\phi_\gamma(t) = \begin{cases}
1 & t \le 0 \\
1-\frac{t}{\gamma} & 0 < t < \gamma \\
0 & t \ge \gamma.
\end{cases}
\end{equation*}
$\phi_\gamma(t) \in [0,1]$ and $\phi_\gamma(\cdot)$ is $1/\gamma$-Lipschitz. The loss function $\ell (f(x), y)$ satisfies:
\begin{equation}\label{eq:ramp_property}
\mathds{1}(y \neq \arg\max_{y' \in [k]} f(x)_{y'}) \le \ell(f(x), y) \le \mathds{1}(f(x)_y \le \gamma + \max_{y' \neq y}f(x)_{y'}).
\end{equation}
$\ell (f(x), y)$ is $2/\gamma$-Lipschitz w.r.t the first argument. Therefore, by replacing $\rho$ by $1/\gamma$ in the right-hand side of the upper bound in Theorem \ref{thm:main}, we obtain a margin bounds for robust generalization of the perdition errors:
{\footnotesize\begin{equation*}
    \begin{aligned}
& \mathbb{P}_{(x, y)\sim\mathcal{D}} \left\{ \exists~x' \in \mathcal{B}(x)~\text{s.t.}~ y \neq \arg\max_{y' \in [k]} f(x')_{y'} \right\} 
-  \frac{1}{n}\sum_{i=1}^n \mathds{1}\bigg( \exists~x_i' \in \mathcal{B}(x_i)~\text{s.t.}~ f( x_i' )_{y_i} \le \gamma + \max_{y' \neq y_i}f( x_i' )_{y'}\bigg).
\end{aligned}
\end{equation*}}

\paragraph{Magnitude of Adversarial Examples.} We consider the magnitude of $\tilde B$ in common settings. For $\ell_p$ attacks, i.e., $\|x-x'\|_p\leq \varepsilon$, we have $\tilde B=\sup\|x'\|_2\leq \|x\|_2+\|x-x'\|_2\leq B +\max\{1,d^{\frac{1}{2}-\frac{1}{p}}\}\varepsilon$. There exists an additional dependency on the dimension-$d$ within the magnitude $\tilde B$. It arises from the discrepancy between $\ell_p$ attacks and the $\ell_2$-norm of training samples. The impact of $d$ is minimal and can be mitigated by rescaling the norms. Furthermore, it is noteworthy that our analysis is capable of providing bounds for a broad range of adversarial attacks, extending beyond merely $\ell_p$ attacks.

\subsection{Comparison with Existing Bounds for DNNs}
The bound proved for two-layer neural networks by \citet{awasthi2020adversarial} is
\begin{equation}
\label{twolayer,1st}
\tilde{\mathcal{O}}\bigg(\frac{\tilde{B}\|W_1\|\|W_2\|A}{\sqrt{n}}\bigg).
\end{equation}
For a general assumption with a Lipschitz activation function and bounded weights,
\[
A=1+\sqrt{d(m+1)}.
\]
The dependency on width and dimension is $\mathcal{O}(\sqrt{md})$, which is larger than that of standard bounds. With an additional assumption for ReLU activation functions and special weights (cf. Theorem 9 in \citet{awasthi2020adversarial}),
\[
A=\mathcal{C}_{\mathcal{S}}^\star\sqrt{\Pi_\mathcal{S}^\star}.
\]
In this setting, the bound is width- and depth-independent.

In our earlier work \citep{xiao2022adversarial}, we provided a bound for DNNs in
\begin{equation}
    \label{xiaobound}
\mathcal{O}\bigg(\frac{\tilde{B}m\sqrt{L\log L}\prod_{i=1}^L\|W_i\|_{op}}{\sqrt{n}}\bigg).
\end{equation}
The dependency on width $\mathcal{O}(m)$ can be further reduced to $\mathcal{O}(\sqrt{rm})$ in a low-rank scenario, where $r$ is the rank of each weight matrix. Even in this scenario, the dependency on width is still increased by $\mathcal{O}(\sqrt{m})$ when compared to the standard bounds. Notably, the bound in Equation \ref{xiaobound} has a lower dependence on $L$, which seems to be a trade-off between depth and width, and the width is at least no smaller than the data dimension. However, a model with good generalization ability should scale with the data dimension. A width-independent bound is more desirable.

The bound proved by \citet{mustafa2022generalization} is
\begin{equation}
    \label{musbound}
\tilde{\mathcal{O}}\bigg(\underbrace{\frac{\tilde{B}L\prod_{i=1}^L\|W_i\|_{op}}{\sqrt{n}}\bigg(\sum_{i=1}^L\frac{\|W_i\|_{2,1}^{2}}{\|W_i\|_{op}^{2}}\bigg)^{1/2}}_{\text{Term 1}}\times\underbrace{\bigg(\tilde{L}_{\text{log}}\bigg)}_{\text{Term 2}\geq\mathcal{O}(\sqrt{Ld})}\bigg),
\end{equation}
where
\begin{equation*}
    \tilde{L}_{\text{log}}=\log^{\frac{1}{2}}\bigg(\bigg(\frac{C_1\tilde{B}\Gamma n}{\gamma}+C_2\bar m\bigg)n\bigg(\frac{6\varepsilon\lambda n}{\gamma}\bigg)^d+1\bigg)\log(n),
\end{equation*}
$\Gamma=\max_{i\in [L]}\prod_{i=1}^L\|W_i\|_{op}\frac{\|W_i\|_{2}m_i}{\|W_i\|_{op}}$, $\bar m =\max_{i\in [L]}m_i $, $\lambda = \frac{2}{\gamma}\prod_{i=2}^L\|W_i\|_{op}\times\|W_1\|_{1,\infty}\sqrt{m_1}$, and $C_1,C_2$ are some constants.

First of all, we consider the term 1 in Eq. (\ref{musbound}). The functional form $(\sum_{i=1}^L(·)^{2/3})^{3/2}$ appearing in \cite{bartlett2017spectrally} may be replaced by the form $L(\sum_{i=1}^L(·)^{2})^{1/2}$ appearing above by using $\|\alpha\|_{2/3} \leq \|\alpha\|_{2}$ which holds for any $\alpha$. Next, we switch our attention to term 2. Since the expression of $\tilde{L}_{\text{log}}$ is rather complicated, we simplify it as
\begin{equation*}
\begin{aligned}
        \tilde{L}_{\text{log}}&=\log^{\frac{1}{2}}\bigg(\bigg(\frac{C_1\tilde{B}\Gamma n}{\gamma}+C_2\bar m\bigg)n\bigg(\frac{6\varepsilon\lambda n}{\gamma}\bigg)^d+1\bigg)\log(n)\\
        &\geq\Omega\bigg(\log^{\frac{1}{2}}\bigg(\frac{6\varepsilon\lambda n}{\gamma}\bigg)^d\bigg)\\
        &\geq\Omega\bigg(\log^{\frac{1}{2}}\bigg(\|W\|_{op}^L\bigg)^d\bigg)\\
        &\geq\Omega(\sqrt{Ld}).
\end{aligned}
\end{equation*}
In conclusion, the bound proposed by \cite{mustafa2022generalization} introduce an additional order (at least) in $\mathcal{O}(\sqrt{Ld})$ and many other factors when comparing to the bound of \cite{bartlett2017spectrally}, v2. 

Notice that our bound reduces to the v1 version of \cite{bartlett2017spectrally}'s bound. Comparing our bound with \cite{mustafa2022generalization}'s bound reduces to comparing $\|W\|_1$ with $\|W\|_{2,1}\tilde{L}_\text{log}$. Since $\|W\|_{2,1}\leq\|W\|_{1}\leq\sqrt{d}\|W\|_{2,1}$, our bound is strictly tighter.


\section{Related Work}
\paragraph{Adversarial Attacks and Defense.} Since 2013, it has been well known that deep neural networks trained by standard gradient descent are highly susceptible to small corruptions to the input data \citep{szegedy2014intriguing,goodfellow2015explaining,chen2017zoo,carlini2017towards,madry2018towards}. One lines of work aimed at increasing the robustness of neural networks \citep{wu2020adversarial,gowal2020uncovering}. Another line of works aimed at finding more powerful attacks \citep{athalye2018obfuscated,tramer2020adaptive,chen2017zoo}.

\paragraph{Robust Generalization.} The work of \citet{schmidt2018adversarially,raghunathan2019adversarial,zhai2019adversarially} has shown that more data can help achieve better robust generalization. The work of \citet{attias2022improved,montasser2019vc} explained generalization in adversarial settings using VC-dimension. \citet{neyshabur2018pac} used a PAC-Bayesian approach to provide a generalization bound for neural networks. The work of \cite{farnia2018generalizable,xiao2023pac} extended the PAC-Bayes analysis to adversarial settings. However, as pointed out in \cite{bartlett2017spectrally}, PAC-bayes bound is not as tight as Rademacher complexity bound. There exist a trade-off between standard and robust accuracy in adversarial training \citep{raghunathan2020understanding,javanmard2020precise,mehrabi2021fundamental,javanmard2022precise,javanmard2023adversarial}. In another line of our research, we study the poor robust generalization through the lens of uniform stability \citep{xiao2022stability,xiao2022adaptive,xiao2022smoothed}. Even though adversarial training helps when enough data is available, it may hurt robust generalization in the small sample size regime \citet{clarysse2022adversarial}.

\paragraph{Rademacher Complexity.} \citet{golowich2018size} introduced an alternative layer peeling technique and obtained a size-independent bound. However, as pointed out in (\citet{telgarsky2021deep}, Sec. 16.2), \citet{golowich2018size}'s Frobenius norm bound is still larger than \citet{bartlett2017spectrally}'s spectral norm bound, which is the best known Rademacher complexity and covering number bound for DNNs in a standard setting.
\section{Conlusion}
This paper introduces upper bounds for ARC that match the upper bounds in standard scenarios \citep{bartlett2017spectrally}. The primary challenge we tackle is the computation of the covering number for adversarial function classes. To address this, we propose a novel concept called the uniform covering number, tailored specifically for adversarial examples. This approach successfully bridges the gap between Rademacher complexity measures in both robust and standard generalization contexts. We believe that the introduced concept of the uniform covering number will be of significant value to the theoretical community. For instance, it has the potential to be adapted for a variety of machine learning problems and algorithms where the training samples are dynamic, not static.
\newpage
\section*{Acknowledgments} We would like to thank all the anonymous reviewers for
their comments and suggestions. This work was supported in part by NIH grants, RF1AG063481 and U01CA274576, NSF DMS-2310679, a Meta Faculty Research Award, and Wharton AI for Business. The content is solely the responsibility of the authors and does not necessarily represent the official views of the NIH.

\bibliography{main.bib}

\newpage
\appendix
\section{Details of Existing ARC Bounds}
In this section, we provide the details of the adversarial bound listed in Table \ref{tab:my_label}.

\subsection{Type (\uppercase\expandafter{\romannumeral1}): Robust Loss and Matching Bounds}
We first state the best-known result in linear cases from \cite{awasthi2020adversarial}.
\begin{theorem}
    [\cite{awasthi2020adversarial}, Theorem 4]
    \label{thm:linear}
    Let $\mathcal{H} := \{y\langle w,x\rangle \mid \|w\|_r \leq W\}$ be be the class of linear functions and $\tilde{\mathcal{H}} := \{\min_{\|x-x'\|_p \leq\varepsilon} y\langle w,x\rangle \mid\|w\|_r\leq W\}$. Then it holds that
\[\max\bigg\{\mathcal{R}(\mathcal{H}),\varepsilon \frac{W\max\{d^{1-\frac{1}{p}-\frac{1}{r}},1\}}{2\sqrt{2n}}\bigg\}\leq\mathcal{R}(\tilde{\mathcal{\mathcal{H}}})\leq\mathcal{R}(\mathcal{H})+\varepsilon \frac{W \max\{d^{1-\frac{1}{p}-\frac{1}{r}},1\}}{2\sqrt{n}}.\]
\end{theorem}
Notice that when the perturbation is measured in $\ell_\infty$-norm, i.e. $p = \infty$, Theorem \ref{thm:linear} recovers the bound of \citet{yin2019rademacher}, and provides a finer analysis of the dependence on the input dimensionality as compared to the recent work of \citet{khim2018adversarial} on linear hypothesis classes. Furthermore, when $\varepsilon=0$, as expected, the ARC equals the standard Rademacher complexity of linear models. 

By setting $\mathcal{R}(\mathcal{H})=\mathcal{O}(BW/\sqrt{n})$, the bound becomes $\mathcal{O}((B+\varepsilon \max\{d^{1-\frac{1}{p}-\frac{1}{r}},1\})W/\sqrt{n})$. Here $B+\varepsilon \max\{d^{1-\frac{1}{p}-\frac{1}{r}},1\}=\tilde B$ represents the magnitude of adversarial examples. Consequently, $\tilde B$ is an unavoidable term in the ARC bounds.
\subsection{Type (\uppercase\expandafter{\romannumeral2}): DNNs and Matching Bounds}
In this section, we introduce the surrogate losses listed in Table \ref{tab:my_label}.

\paragraph{Tree Transformation Loss.}  The work of \citep{khim2018adversarial} introduced a tree transformation $T$ and showed that $\max_{\|x-x'\|\leq \epsilon}\ell(f( x),y)\leq\ell(Tf( x),y)$. The tree transformation pushes the maximization through each
layer, thus multiplying the bound slack. Then, we have the following upper bound for the adversarial population risk. For $\delta\in(0,1)$,
\begin{eqnarray*}
\tilde{R}_{\mathcal{D}}(f)\leq R_{\mathcal{D}}(Tf)\leq R_{\mathcal{S}}(Tf)+2\rho\mathcal{R}(T\circ\mathcal{F}_{|\mathcal{S}})+3\sqrt{\frac{\log\frac{2}{\delta}}{2n}}.
\end{eqnarray*}
It gives an upper bound of the robust population risk by the empirical risk and the standard Rademac-her complexity of $T\circ f$, i.e., $\mathcal{R}(T\circ\mathcal{F}_{|\mathcal{S}})$. However, the empirical risk $R_{\mathcal{S}}(Tf)$ in the right-hand side is not the objective in practice. This analysis does not provide a bound for robust generalization gaps.

\paragraph{SDP Relaxation Surrogate Loss.}
In the work of \citep{yin2019rademacher}, the authors defined the SDP surrogate loss as
\begin{eqnarray*}
\hat{\ell}(f( x),y)=\phi_\gamma\bigg(M(f( x),y)-\frac{\epsilon}{2}\max_{k\in [K],z=\pm 1}\max_{P\succeq0, diag(P)\leq 1}\langle zQ(w_{2,k},W_1),P\rangle\bigg)
\end{eqnarray*}
to approximate the adversarial loss for two-layer neural nets. Therefore, the ARC is approximated by the Rademacher complexity on this loss function. The weakness of this approach is the same as that of the previous one.

\paragraph{FGSM Attack Loss.} 
The work of \citep{gao2021theoretical} also considerd the Rademacher complexity in adversarial settings. To deal with the $\max$ operation in the adversarial loss, they consider FGSM adversarial examples. By some assumptions on the gradient, they provide an upper bound for Rademacher complexity on the loss $\ell(f( x_{FGSM}),y)$. They further assumed that the gradient $\|\nabla \ell(f( x),y)\|\geq\kappa $ for all $x$ in the domain. This is a strong assumption and the additional parameter $\kappa$ is in the divider in the final bound. The bound is not controllable when $\kappa\rightarrow 0$. Similar to the tree-transformation loss and SDP-relaxation loss, this approach cannot provide a bound for robust generalization gap.

\section{Proof of the Technical Results}
\subsection{Proof of Lemma \ref{lemma:unupperbound}}
First recall the Maurey sparsification lemma.
\begin{lemma}[Maurey, cf. \citep{pisier1981remarques}]\label{lemma:maurey}
    Fixed a Hilbert space $\mathcal{H}$ with norm $\|\cdot\|$, Let $u \in \mathcal{H}$ be given with representation $u =\sum_{j=1}^d \alpha_j v_j$ where $v_j \in \mathcal{H}$, $\|v_j\|\leq b$, $\alpha_j\geq 0$ and $\alpha=\sum_{j=1}^d \alpha_j\leq 1$. Then for any positive integer $k$, there exists a choice of nonnegative integers $(k_1,\cdots, k_d),
    \sum_{j=1}^d k_i = k$, such that
    \[\bigg\|u-\frac{1}{k}\sum_{j=1}^d k_j v_j\bigg\|^2\leq\frac{\alpha b^2-\|u\|^2}{k}.\]
\end{lemma}
Then, we move to the proof of Lemma \ref{lemma:unupperbound}. Set $N:= 2dm$, $k=\lceil\frac{a^2b^2}{\epsilon^2}\rceil$, and define
\begin{equation*}
    \{V_1,\cdots,V_N\}=\{g e_i e_j: i\in\{1,\cdots,m\}, j\in\{1,\cdots,d\},g\in\{-1,+1\}\},
\end{equation*}
\begin{equation*}
    \mathcal{C}=\bigg\{\frac{a}{k}\sum_{i=1}^N k_iV_i\mid k_i\geq 0,\sum_{i=1}^N k_i=k\bigg\}.
\end{equation*}
For all $X$ such that $\|X\|_2\leq b$ and $W\in\mathcal{W}$,
\[
WX=\sum_{i=1}^m\sum_{j=1}^d W_{i,j}e_i e_j X=a\sum_{i=1}^m\sum_{j=1}^d \frac{W_{i,j}}{a}e_i e_j X\in a \cdot \text{conv}\{V_1 X,\cdots,V_N X\}.
\]
Additionally, $\|V_i X\|_2\leq\|X\|_2\leq b$.
Then, by Lemma \ref{lemma:maurey}, we have
\[
\bigg\|WX-\frac{a}{k}\sum_{i=1}^N k_iV_i X\bigg\|^2\leq\frac{a ^2b^2}{k}\leq\epsilon^2.
\]

Therefore, $\{W'X:W'\in\mathcal{C}\}$ is always an $\epsilon$-cover of $\{WX:W\in\mathcal{W}\}$, for all $X\in\mathcal{X}=\{X \mid \|X\|_2\leq b\}$. By the definition of $\epsilon$-uniform cover, $\mathcal{C}$ is the desired $\epsilon$-uniform cover of $\mathcal{W}$. The uniform covering number is $|\mathcal{C}|=(2dm)^k=(2dm)^{\lceil\frac{a^2b^2}{\epsilon^2}\rceil}$. Thus we complete the proof that
\begin{equation*}
    \ln \mathcal{UN}_{\mathcal{X}}(\mathcal{W},\epsilon,\|\cdot\|) \leq\bigg\lceil\frac{a^2b^2}{\epsilon^2}\bigg\rceil \ln(2dm).
\end{equation*}

\subsection{Proof of Theorem \ref{thm:main2}}
First of all, we define
\[
\epsilon_i=\frac{\epsilon}{\rho\rho_i\prod_{j=1}^{i-1} \rho_j s_j}\bigg(\frac{a_i}{s_i}\bigg)^{\frac{2}{3}}\bigg(\frac{1}{\sum_{j=1}^L(a_j/s_j)^{\frac{2}{3}}}\bigg).
\]
Based on the $\rho$-Lipschitz properties of the activation function and the inequality $$\|Wx\|_2\leq\|W\|_\sigma\|x\|_2,$$ 
for all $X_i\in\mathcal{X}_{i}$,
$$\|X_i\|\leq\tilde B \prod_{j=1}^i\rho_j\|W_j\|_\sigma\leq\tilde B \prod_{j=1}^i\rho_j s_j.$$
By, Lemma \ref{lemma:unupperbound},
\begin{equation*}
    \ln \mathcal{UN}_{\mathcal{X}_{i-1}}(\mathcal{W}_i,\epsilon_i,\|\cdot\|) \leq\bigg\lceil\frac{a_i^2(\tilde B \prod_{j=1}^{i-1}\rho_j s_j)^2}{\epsilon_i^2}\bigg\rceil \ln(2m_i m_{i-1}).
\end{equation*}
Then
\begin{equation*}
\begin{aligned}
    \ln \mathcal{N}(\tilde{\mathcal{H}}_{|\mathcal{S}},\epsilon,\|\cdot\|_2)&\leq\sum_{i=1}^L \ln \mathcal{UN}_{\mathcal{X}_{i-1}}(\mathcal{W}_i,\epsilon,\|\cdot\|)\\
    &\leq\sum_{i=1}^L\bigg\lceil\frac{a_i^2(\tilde B \prod_{j=1}^{i-1}\rho_j s_j)^2}{\epsilon_i^2}\bigg\rceil \ln(2m_i m_{i-1})\\
&\leq\frac{\tilde{B}^2\rho\ln(2\bar m^2)}{\epsilon^2}\bigg(\prod_{i=1}^L\rho_i s_i\bigg)\bigg(\sum_{i=1}^L\frac{a_i^{2/3}}{s_i^{2/3}}\bigg)^{3/2}.
\end{aligned}
\end{equation*}
\end{document}